\documentclass[letterpaper]{article} 
\usepackage{aaai2026}  
\usepackage{times}  
\usepackage{helvet}  
\usepackage{courier}  
\usepackage[hyphens]{url}  
\usepackage{graphicx} 
\usepackage{natbib}  
\usepackage{caption} 
\usepackage{algorithm}
\usepackage{algorithmic}

\usepackage{amsmath}
\usepackage{amssymb}
\usepackage{graphicx}
\usepackage{booktabs}
\usepackage{multirow}
\usepackage{multicol}
\newcommand{\se}[1]{{\scriptstyle \pm #1}}

\usepackage{newfloat}
\usepackage{listings}
\DeclareCaptionStyle{ruled}{labelfont=normalfont,labelsep=colon,strut=off} 
\floatstyle{ruled}
\newfloat{listing}{tb}{lst}{}
\floatname{listing}{Listing}
\title{
GammaZero: Learning to Guide Belief-Space Search for Long-Horizon POMDPs with Generalizable Graph Representations
}
\author{
    Rajesh Mangannavar, 
    Prasad Tadepalli 
}
\affiliations{
    Oregon State University, 
Corvallis, OR 97330, USA \\
    \{mangannr,  prasad.tadepalli\}@oregonstate.edu
}

\begin{document}

\maketitle

\begin{abstract}
We introduce an uncertainty-aware graph representation framework for learning to guide planning in Partially Observable Markov Decision Processes (POMDPs). Unlike existing approaches that require domain or problem size specific neural architectures, GammaZero leverages a unified graph-based belief representation that enables generalization across problem sizes within a domain. Our key insight is that belief states can be systematically transformed into uncertainty-aware graphs where structural patterns learned on small problems transfer to larger instances. We employ a graph neural network with a decoder architecture to learn value functions and policies from expert demonstrations on computationally tractable problems, then apply these learned heuristics to guide Monte Carlo tree search on larger problems. Experimental results on standard POMDP benchmarks demonstrate that GammaZero achieves comparable performance to BetaZero when trained and tested on the same-sized problems, while enabling zero-shot generalization to problems 2-6× larger than those seen during training. 



\end{abstract}

\section{Introduction}

Partially Observable Markov Decision Processes (POMDPs) provide a principled framework for sequential decision-making under uncertainty, where agents must act based on incomplete information about the true state of the environment~\cite{kaelbling1998planning}. This partial observability arises naturally in many real-world applications, from autonomous driving where sensors provide limited field-of-view~\cite{hoel2019combining}, to robotic manipulation where object properties must be inferred through interaction~\cite{lauri2022partially}, to subsurface exploration where underground structures can only be observed at sparse drilling locations~\cite{mern2023intelligent}. The ability to reason explicitly about uncertainty while planning makes POMDPs particularly well-suited for safety-critical applications where robust decision-making is essential.

Despite their advantages, solving POMDPs exactly becomes computationally intractable for all but the smallest problems due to the curse of dimensionality in belief space~\cite{shani2013survey}. Online planning algorithms such as POMCP~\cite{silver2010pomcp} and POMCPOW~\cite{sunberg2018online} have made significant progress by using Monte Carlo tree search to focus computational effort on reachable belief states. However, these methods face fundamental limitations when dealing with long planning horizons and high-dimensional state spaces. Without effective heuristics to guide search,
online planners struggle to look sufficiently far ahead to discover rewarding action sequences that may require extended information gathering~\cite{ye2017despot}.

The success of AlphaZero in fully observable domains demonstrates that learned approximations can effectively replace hand-crafted heuristics~\cite{silver2018general}. Algorithms like BetaZero~\cite{betazero2024} have extended this to partially observable domains by training neural networks to predict values and policies from belief states. However, reliance on fixed-size inputs creates a representational bottleneck, preventing generalization to larger problem sizes.  

Graph Neural Networks (GNNs) offer a powerful alternative to address this scalability gap. While prior work has successfully leveraged GNNs to generalize policies in fully observable planning domains~\cite{mangannavargraph}, their application to the probabilistic belief spaces of POMDPs remains largely unexplored. 
In this work, we present GammaZero, a novel framework for learning to guide belief space search in POMDPs using uncertainty-aware graph representations, enabling a model trained on computationally tractable "toy" problems to generalize to large-scale instances that are computationally expensive for traditional online planners.  
Our contributions are:

\begin{enumerate}
\item \textbf{A graph-based belief representation for POMDPs} that transforms belief states into action-centric graphs encoding relationships between objects, their attributes, and actions. This enables learning from small problems to generalize to larger instances.


\item \textbf{Experimental validation:} Integration of learned value function and policy with MCTS showing GammaZero achieves comparable performance to BetaZero on same-sized problems, while enabling zero-shot generalization to problems 2-6× larger than training instances.
\end{enumerate}

These contributions address long-horizon planning under uncertainty by combining graph neural networks' generalization capabilities with POMDP belief-space search.

\section{Related Work}

\textbf{Monte Carlo Tree Search:} 
Monte Carlo Tree Search (MCTS) is a best-first search algorithm that builds a search tree incrementally through repeated simulations \cite{browne2012survey}. 
Each iteration consists of four phases: traversing the tree using a selection policy, adding new nodes to the leafs, simulating a rollout policy until terminal states, and updating statistics along the traversed path. The UCB1 formula commonly guides selection by balancing average rewards with exploration bonuses based on visit counts \cite{kocsis2006bandit}. While MCTS has achieved remarkable success in games and planning \cite{silver2016mastering,silver2017mastering,silver2018general}, applying it to POMDPs requires careful handling of belief states and typically demands substantial computational resources for reliable value estimates.

\textbf{Online POMDP Planning:} Classical online POMDP planning algorithms employ tree search methods to determine optimal actions through forward simulation. POMCP \cite{silver2010pomcp} extends Monte Carlo tree search to POMDPs by maintaining particle beliefs at tree nodes, enabling planning in large state spaces. POMCPOW \cite{sunberg2018online} further extends this to continuous observation spaces through progressive widening. DESPOT \cite{ye2017despot} regularizes the search tree to focus on high-probability scenarios, while AdaOPS \cite{wu2021adaptive} adaptively adjusts particle beliefs to maintain value function bounds. While these methods rely heavily on domain-specific heuristics for value estimation and action selection,  
our approach learns generalizable graph-based representations that eliminate the need for hand-crafted heuristics. 

\textbf{Learning for Online POMDP Planning:} Recent work has explored combining offline learning with online planning to reduce reliance on heuristics. BetaZero \cite{betazero2024} and ConstrainedZero \cite{constrained_zero} learn neural network approximations of optimal policies and value functions offline, then use these to guide online MCTS. LeTS-Drive \cite{cai2022closing} similarly combines offline learning with online HyP-DESPOT planning for autonomous driving domains. These approaches fundamentally rely on fixed-size belief representations that must be predetermined for each domain. Our work differs by introducing a graph-based belief representation that naturally handles variable-sized problems and explicitly captures action-state relationships, enabling zero-shot generalization to problems significantly larger than those in training.

\textbf{Learning for Classical Planning :} The planning community has developed several approaches for learning generalized policies that transfer across problem sizes. GPL \cite{generalized_planning_deep_rl} learns value functions over relational state representations using GNNs, selecting actions by evaluating successor states. However, maintaining globally consistent value estimates becomes increasingly challenging as problems scale. ASNets \cite{toyer2020asnets} employs alternating action and proposition layers with weight sharing, but its fixed-depth architecture limits reasoning about long dependency chains. GRAPL \cite{chrestien_optimize} learns to rank actions using canonical abstractions but lacks explicit modeling of action-object relationships and parameter dependencies.

\textbf{Graph Representations for Planning:} Graph neural networks have shown promise for learning generalizable planning policies due to their ability to handle variable-sized inputs and capture relational structure \cite{gnn_base,hamilton2020graph}. Existing work primarily focuses on deterministic, fully observable domains where states map directly to graph structures \cite{strips_hgn,karia2021learning}. Most recently, GABAR \cite{mangannavargraph} introduces an action-centric graph representation for deterministic planning, using a GNN encoder and a GRU-based decoder to incrementally construct grounded actions. While GABAR demonstrates strong generalization in fully observable domains, it cannot handle the belief uncertainty inherent in POMDPs. 
The key challenge addressed in the current paper is extending the graph representations to POMDPs by encoding belief uncertainty while preserving the structural patterns that enable generalization.

\section{Problem Formulation}

We formulate the problem of learning to guide belief space search as a supervised learning task. Given a POMDP $\mathcal{M} = \langle S, A, T, R, O, Z, \gamma, b_0 \rangle$ with states $S$, actions $A$, transition function $T$, reward function $R$, observations $O$, observation function $Z$, discount factor $\gamma$, and initial belief $b_0$, our goal is to learn both a value function $V_\theta: \mathcal{B} \rightarrow \mathbb{R}$ and a policy $\pi_\phi: \mathcal{B} \times A \rightarrow [0,1]$ over the belief space $\mathcal{B}$ \cite{kaelbling1998planning}.

The training data consists of tuples $(b_i, v_i^*, \pi_i^*)$ where $b_i$ is a belief state encountered during expert planning, $v_i^*$ is the optimal value-to-go from that belief, and $\pi_i^*$ is the optimal action distribution. These targets are obtained by running optimal or near-optimal planners on small problem instances where exact solutions are computationally feasible. The learned value function and policy are then used to guide online search on larger problem instances.

\section{GammaZero}

\begin{figure*}[t]
\centering
\includegraphics[width=1.0\textwidth]
{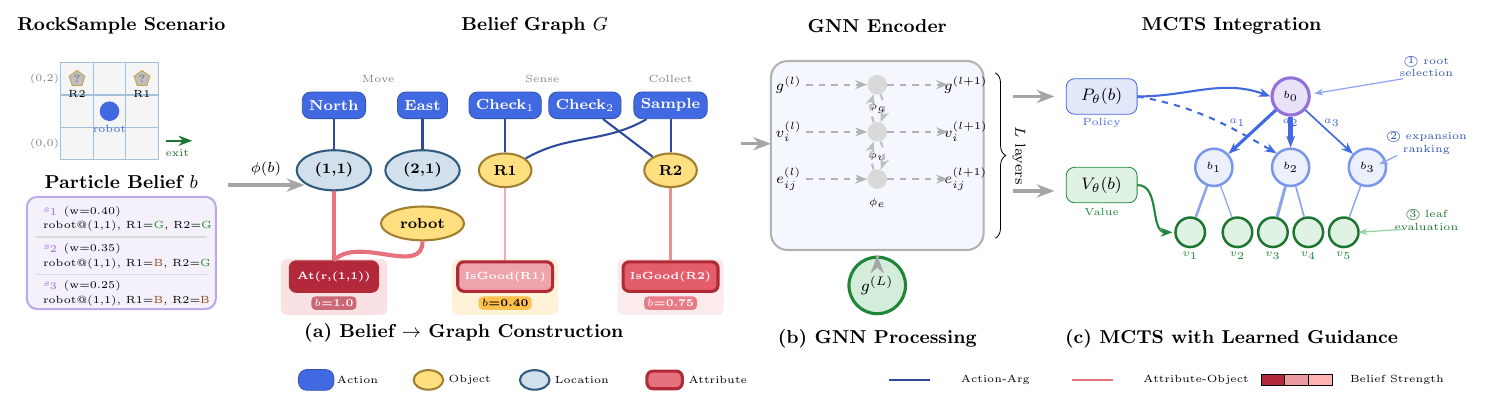}
\caption{GammaZero framework. (a) Particle beliefs are transformed into uncertainty-aware graphs encoding objects, their attributes and actions. (b) A GNN processes the graph through $L$ message-passing layers, updating node, edge, and global features. (c) The learned policy $P_\theta$ guides MCTS action selection while $V_\theta$ evaluates leaf nodes, replacing expensive rollouts.}
\label{fig:gz_overview}
\end{figure*}

GammaZero operates in two phases: an offline learning phase where we train graph neural network approximations from expert-generated data on small problem instances, and an online planning phase where these learned models guide MCTS in belief space. The key innovation is our graph-based belief representation that captures relationships between actions, objects, and their associated uncertainties, enabling zero-shot generalization to larger problem instances without retraining. Figure~\ref{fig:gz_overview} provides an overview of the framework. We detail each component below.

\begin{algorithm}[t]
\caption{GammaZero MCTS Planning}
\label{alg:mcts_planning}
\begin{algorithmic}[1]
\REQUIRE Belief $b$, GNN $f_\theta = (V_\theta, P_\theta)$, simulations $n_{\text{sim}}$, depth $d$
\ENSURE Selected action $a^*$
\STATE $\mathcal{T} \gets \textsc{InitTree}(b)$ \label{line:init_tree}
\STATE $G \gets \phi(b)$ \label{line:belief_to_graph} \COMMENT{Convert belief to graph}
\FOR{$i = 1$ to $n_{\text{sim}}$} \label{line:sim_loop_start}
    \STATE \textsc{Simulate}$(\mathcal{T}, b, G, d)$
\ENDFOR \label{line:sim_loop_end}
\FOR{each action $a \in \textsc{Children}(\mathcal{T}.\text{root})$} \label{line:action_select_start}
    \STATE $\pi(a) \propto N(b, a)^{z_n} \cdot \exp(z_q \cdot Q(b, a))$
\ENDFOR
\STATE $a^* \gets \arg\max_a \pi(a)$ \label{line:action_select_end}
\RETURN $a^*$
\end{algorithmic}
\end{algorithm}

\begin{algorithm}[t]
\caption{GammaZero MCTS Simulation}
\label{alg:mcts_simulate}
\begin{algorithmic}[1]
\REQUIRE Tree $\mathcal{T}$, belief $b$, graph $G$, depth $d$
\ENSURE Value estimate $q$
\IF{$d = 0$ or \textsc{IsTerminal}$(b)$}
    \RETURN $\mathbf{V_\theta(G)}$ \label{line:leaf_eval_terminal} \COMMENT{Leaf evaluation}
\ENDIF
\IF{\textsc{IsLeaf}$(\mathcal{T}, b)$}
    \STATE \textsc{Expand}$(\mathcal{T}, b, \mathbf{P_\theta(G)})$ \label{line:policy_expand} \COMMENT{Policy-guided expansion}
    \RETURN $\mathbf{V_\theta(G)}$ \label{line:leaf_eval_expand}
\ENDIF
\STATE $a \gets \arg\max_a \left[ Q(b, a) + c \cdot \mathbf{P_\theta(a \mid G)} \cdot \frac{\sqrt{\sum_{a'} N(b, a')}}{1 + N(b, a)} \right]$ \label{line:ucb_select}
\STATE $s \sim b$ \label{line:sample_state}
\STATE $s', o, r \gets \textsc{Step}(s, a)$ \label{line:sim_step}
\STATE $b' \gets \textsc{UpdateBelief}(b, a, o)$ \label{line:update_belief}
\IF{$b' \notin \mathcal{T}$} \label{line:check_new_node}
    \STATE \textsc{AddNode}$(\mathcal{T}, b, a, b')$ \label{line:add_node}
\ENDIF
\STATE $G' \gets \phi(b')$ \label{line:convert_graph}
\STATE $q \gets r + \gamma \cdot \textsc{Simulate}(\mathcal{T}, b', G', d - 1)$ \label{line:recursive_sim}
\STATE $N(b, a) \gets N(b, a) + 1$ \label{line:update_n}
\STATE $Q(b, a) \gets Q(b, a) + \frac{q - Q(b, a)}{N(b, a)}$ \label{line:update_q}
\RETURN $q$
\end{algorithmic}
\end{algorithm}

\subsection{Belief State as Uncertainty-Aware Graph}

\subsubsection{Graph Construction Principles}

Our graph representation transforms particle-based belief states into structured graphs that capture both the uncertainty inherent in partial observability and the action-centric relationships necessary for decision-making. The core principle is to create a sparse yet expressive representation where the graph topology itself encodes belief uncertainty, attribute instance nodes exist only when sufficient particle support justifies their inclusion, naturally encoding the multimodal nature of beliefs through the presence and absence of nodes. The graph schema, i.e., the object types, attribute types, and action types, is defined once per domain; given this schema, graphs are constructed automatically from any belief state. 

\begin{figure*}[t]
\centering
\includegraphics[width=0.9\textwidth]{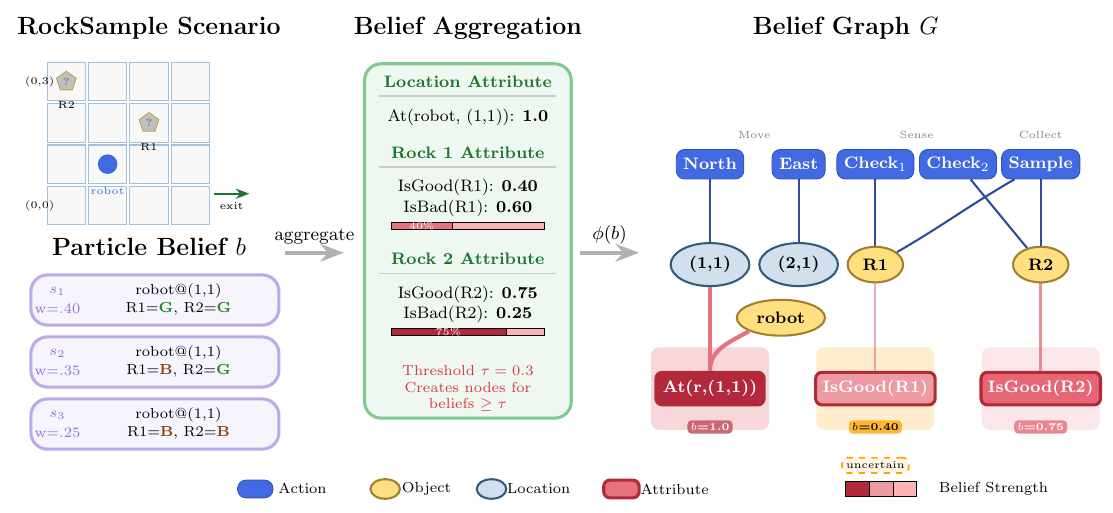}
\caption{Belief graph construction pipeline. A particle belief from RockSample is aggregated into attribute probabilities (e.g., IsGood(R1)$=0.40$, IsGood(R2)$=0.75$), with attributes exceeding threshold $\tau$ instantiated as nodes. The resulting graph encodes objects, actions, and belief-weighted attributes, where edge thickness and node color intensity reflect uncertainty. Note: locations are visually distinguished for clarity but are treated as object nodes in $V_{\text{obj}}$.}

\label{fig:graph_construction}
\end{figure*}

Given a belief state $b$ represented as a set of weighted particles $\{(s_i, w_i)\}_{i=1}^n$, where each particle $s_i$ is a complete world state assigning values to all object attributes, we construct a graph $G = (V, E, X^V, X^E)$ where $V$ is the node set, $E$ is the edge set, $X^V$ maps nodes to feature vectors, and $X^E$ maps edges to feature vectors. We assume an object-centric state representation: each state decomposes into a fixed set of objects, each with typed attributes (e.g., \texttt{At(robot)}, \texttt{IsGood(rock1)}). To build the graph, we compute the probability of each attribute-value pair by aggregating particle weights where that assignment holds. This graph representation approximates the joint belief through independent per-attribute distributions; inter-attribute correlations (e.g., ``all objects are co-located'') present in the particles are not preserved in the graph (see Limitations). Figure~\ref{fig:graph_construction} illustrates this construction on a RockSample domain, showing how particle beliefs are aggregated into per-attribute probabilities and selectively instantiated as graph nodes based on the threshold $\tau$.

The node set consists of four distinct types:
\begin{equation}
V = V_{\text{obj}} \cup V_{\text{attr}} \cup V_{\text{act}} \cup \{v_{\text{global}}\}
\end{equation}

\noindent where:
\begin{itemize}
\item $V_{\text{obj}}$ contains \textbf{object nodes} representing all distinct entities in the environment. This unifies movable entities (e.g., robot, box, package) and spatial entities (e.g., rooms, hallways, distinct waypoints) into a single set. These nodes persist across all beliefs and serve as the valid arguments over which actions and attributes are defined. They encode entity-specific properties, such as type (e.g., \texttt{is\_location}, \texttt{is\_item}) and static attributes.

\item $V_{\text{attr}}$ contains \textbf{attribute instance nodes} representing object attributes that hold with sufficient probability in the belief. Each attribute node encodes a specific attribute-value assignment for an object. For example, \texttt{At(robot)= (kitchen)} represents that the \texttt{At} attribute of \texttt{robot} has value \texttt{kitchen}. These nodes are created selectively: an attribute instance is instantiated only when $\sum_i w_i \cdot \mathbf{1}[\text{attr}(obj) = val \text{ in } s_i] \geq \tau$. This selective instantiation allows the graph structure to directly encode which attribute-value assignments are plausible under the current belief.

\item $V_{\text{act}}$ contains \textbf{action nodes} representing parameterized actions available in the domain (e.g., \texttt{move(?from, ?to)}, \texttt{pick(?object)}). These nodes enable the model to reason about action applicability by examining their connections to the relevant entities in $V_{\text{obj}}$ and condition nodes in $V_{\text{attr}}$.

\item $v_{\text{global}}$ is a \textbf{global aggregation node} that maintains a holistic representation of the belief state and propagates information across distant nodes in the graph.
\end{itemize}

This structure serves multiple purposes. First, it separates persistent structural elements (entities and actions) from belief-dependent elements (attribute instances), allowing the model to distinguish between static domain knowledge and dynamic uncertainty. Second, the unification of locations and objects into $V_{\text{obj}}$ simplifies the topology, treating spatial navigation and object manipulation as fundamentally similar operations defined by relationships between entities. Third, the selective creation of attribute nodes naturally handles multimodal distributions - if the belief assigns significant probability to the robot being in either the kitchen or hallway, both \texttt{At(robot, kitchen)} and \texttt{At(robot, hallway)} nodes will exist, with edge weights encoding their respective probabilities.

\subsubsection{Belief-Driven Sparsity}

A key innovation is the belief-driven creation of attribute instance nodes. Rather than instantiating all possible attribute groundings, we create nodes only when the aggregated particle support exceeds a threshold $\tau$:
\begin{equation}
\text{CreateNode}(\text{attr}(args)) \text{ iff } 
\sum_{i=1}^n w_i \cdot \mathbf{1}[\text{attr}(args) \in s_i] \geq \tau
\end{equation}

This threshold-based construction serves multiple purposes: it naturally encodes the belief distribution through the graph topology (existence implies plausibility), reduces computational complexity by avoiding nodes for unlikely hypotheses, and enables the model to learn patterns from structural presence/absence rather than just numerical features.

\subsubsection{Edge Construction and Belief Encoding}

Edges in our graph serve as the primary mechanism for encoding relationships between entities and, crucially, how belief uncertainty affects these relationships. The edge set consists of three primary categories:

\begin{equation}
E = E_{\text{attr-obj}} \cup E_{\text{act-obj}} \cup E_{\text{attr-act}}
\end{equation}

\noindent where:
\begin{itemize}

\item $E_{\text{attr-obj}}$ contains \textbf{attribute-object edges} connecting attribute instance nodes to the relevant object nodes. Each attribute node connects to the object it describes and to the node representing its value. For instance, the node \texttt{At(robot) = (kitchen)} has edges to \texttt{robot} (the object whose \texttt{At} attribute is described) and to \texttt{kitchen} (the value of the attribute). Edge features distinguish between these two roles, and edge weights encode the belief probability of this attribute-value assignment.

\item $E_{\text{act-obj}}$ contains \textbf{action-object edges} linking actions to objects that can serve as their parameters. These edges capture action applicability constraints and expected outcomes. For example, an edge from \texttt{check(?rock)} to \texttt{rock\_3} encodes not only that rock\_3 can be checked, but also the observation accuracy (decreasing with distance) and expected information gain.

\item $E_{\text{attr-act}}$ contains \textbf{attribute-action edges} connecting attribute instances to actions that either require them as preconditions or produce them as effects. These edges encode how current beliefs constrain future actions and their expected outcomes.
\end{itemize}

Edge features capture multiple layers of information:

\begin{equation}
X^E_{ij} = [\phi_{\text{type}}(e_{ij}), \phi_{\text{role}}(e_{ij}), \phi_{\text{belief}}(e_{ij}), \phi_{\text{support}}(e_{ij})]
\end{equation}

\noindent where:
\begin{itemize}
\item $\phi_{\text{type}}(e_{ij}) \in \{0,1\}^{10}$ is a 10-dimensional one-hot vector encoding the edge type (action-object, object-action, attribute-object, object-attribute, attribute-action, action-attribute, etc.), with bidirectional edges having distinct types to capture directionality.

\item $\phi_{\text{role}}(e_{ij}) \in \{0,1\}^2$ is a 2-dimensional one-hot vector encoding the edge role for attribute edges: whether the edge connects to the object whose attribute is described (owner) or to the node representing the attribute's value.

\item $\phi_{\text{belief}}(e_{ij}) \in [0,1]$ is the continuous belief probability for this relationship, computed as $\sum_{i=1}^n w_i \cdot \mathbf{1}[\text{relationship holds in } s_i]$.

\item $\phi_{\text{support}}(e_{ij})$ is a categorical discretization of $\phi_{\text{belief}}$ into consensus levels: \textit{unanimous} ($>95\%$), \textit{strong} ($70$--$95\%$), \textit{weak} ($30$--$70\%$), or \textit{split} ($<30\%$). Both features derive from the same aggregated particle support, but the categorical version serves as an inductive bias that provides the network with explicit signal for patterns such as ``split support suggests information-gathering actions are valuable.'' For example, if $\phi_{\text{belief}} = 0.72$ for IsGood(rock1), then $\phi_{\text{support}}$ is \textit{strong}; a nearby value of $0.68$ would instead be \textit{weak}, making the transition salient even though the continuous values are close. A sufficiently expressive GNN could learn such boundaries from $\phi_{\text{belief}}$ alone, but the categorical feature makes these patterns easier to learn from limited training data.
\end{itemize}

This rich edge representation serves three critical functions. First, it encodes \textbf{belief-weighted relationships}---the edge weight between an attribute and its value directly represents how strongly this relationship is believed to hold. Second, it captures \textbf{action-specific context}---edges from check actions to objects encode observation accuracy based on distance, while edges from sample actions encode co-location requirements. Third, it provides \textbf{particle support metadata} that distinguishes between confident beliefs (high support from many particles) and uncertain hypotheses (support from few particles), enabling the model to reason about both the magnitude and confidence of beliefs.

This graph construction applies broadly to object-centric POMDPs domains where the state can be decomposed into discrete entities with attributes and relational structure between them (e.g., spatial adjacency, prerequisite dependencies). The model learns to interpret structural patterns (e.g., ``high entropy on attribute nodes connected to an action indicates information-gathering value'') rather than domain-specific features, enabling transfer learning across problem instances of vastly different scales within the same domain.

\subsection{Graph Neural Network Architecture}

We employ a graph neural network that processes the belief graph through $L$ rounds of message passing:

\begin{align}
e_{ij}^{(l+1)} &= \phi_e([e_{ij}^{(l)}, v_i^{(l)}, v_j^{(l)}, g^{(l)}]), \\
v_i^{(l+1)} &= \phi_v([v_i^{(l)}, \text{AGG}(\{e_{ij}^{(l+1)} : j \in \mathcal{N}(i)\}), g^{(l)}]), \\
g^{(l+1)} &= \phi_g([g^{(l)}, \text{AGG}(\{v_i^{(l+1)}\}), \text{AGG}(\{e_{ij}^{(l+1)}\})])
\end{align}

where $\phi_e$, $\phi_v$, and $\phi_g$ are learned update functions, and AGG represents an attention-weighted aggregation. The global node $g$ enables rapid information propagation across the graph, crucial for maintaining performance as problem size increases.

The network outputs both a value estimate and action probabilities:

\begin{align}
V_\theta(G) &= \text{MLP}_v(g^{(L)}), \\
P_\theta(a|G) &= \text{softmax}(\text{MLP}_p([g^{(L)}, v_a^{(L)}])).
\end{align}

\subsection{Data Collection and Training}

\subsubsection{Expert Data Generation}

\begin{algorithm}[t]
\caption{Collect Expert Trajectory}
\label{alg:collect_trajectory}
\begin{algorithmic}[1]
\REQUIRE POMDP $\mathcal{M}$, expert planner $\pi^*$, initial belief $b_0$, max steps $T$
\ENSURE Dataset $\mathcal{D}$ of (belief, value, action) tuples
\STATE $\mathcal{D} \gets \emptyset$, $b \gets b_0$, $t \gets 0$
\STATE $\mathcal{H} \gets [(b_0, \cdot, \cdot)]$ \COMMENT{History buffer for return computation}
\WHILE{$t < T$ and not $\textsc{IsTerminal}(b)$}
    \STATE $a^*, Q^* \gets \pi^*(b)$ \label{line:expert_query} \COMMENT{Expert action and Q-values}
    \STATE $\mathcal{H}[t].\text{action} \gets a^*$
    \STATE $s \sim b$ \COMMENT{Sample state from belief}
    \STATE $s', o, r \gets \textsc{Step}(\mathcal{M}, s, a^*)$ \label{line:env_step}
    \STATE $\mathcal{H}[t].\text{reward} \gets r$
    \STATE $b \gets \textsc{UpdateBelief}(b, a^*, o)$ \label{line:belief_update}
    \STATE $\mathcal{H}.\text{append}((b, \cdot, \cdot))$
    \STATE $t \gets t + 1$
\ENDWHILE
\STATE \COMMENT{Compute discounted returns from end of episode}
\STATE $G \gets 0$ \label{line:return_start}
\FOR{$\tau = t$ down to $0$}
    \STATE $G \gets \mathcal{H}[\tau].\text{reward} + \gamma \cdot G$
    \STATE $\mathcal{D} \gets \mathcal{D} \cup \{(\mathcal{H}[\tau].\text{belief}, G, \mathcal{H}[\tau].\text{action})\}$ \label{line:return_end}
\ENDFOR
\RETURN $\mathcal{D}$
\end{algorithmic}
\end{algorithm}

We collect training data by running optimal or near-optimal planners on small problem instances where exact solutions are computationally feasible. Algorithm~\ref{alg:collect_trajectory} details our trajectory collection procedure. For each belief state encountered, we query the expert planner for the optimal action and Q-values (line~\ref{line:expert_query}), execute the action in the environment (line~\ref{line:env_step}), and update the belief based on the received observation (line~\ref{line:belief_update}). After the episode terminates, we compute discounted returns via backward induction (lines~\ref{line:return_start}--\ref{line:return_end}), associating each visited belief with its value-to-go. This supervised approach leverages existing planning algorithms to generate high-quality training targets.

\subsubsection{Loss Functions}

The network is trained using a combination of mean squared error for value prediction and cross-entropy for action classification:

\begin{equation}
\mathcal{L} = \lambda_v \|V_\theta(G) - v^*\|^2 + \lambda_p \mathcal{L}_{\text{CE}}(P_\theta(\cdot|G), a^*)
\end{equation}

where $\mathcal{L}_{\text{CE}}$ is the cross-entropy loss and $\lambda_v, \lambda_p$ are weighting coefficients. We use a shared GNN encoder with separate MLP heads rather than independent networks, as this avoids redundant feature learning and halves the inference cost during MCTS.

\subsection{Integration with Monte Carlo Tree Search}

During online execution, the learned approximations enhance MCTS as detailed in Algorithms~\ref{alg:mcts_planning} and~\ref{alg:mcts_simulate}. Algorithm~\ref{alg:mcts_planning} initializes the search tree and converts the current belief to a graph representation (lines~\ref{line:init_tree}--\ref{line:belief_to_graph}), then iteratively runs simulations to build the tree (lines~\ref{line:sim_loop_start}--\ref{line:sim_loop_end}). After all simulations complete, action selection combines visit counts and Q-values (lines~\ref{line:action_select_start}--\ref{line:action_select_end}).

Algorithm~\ref{alg:mcts_simulate} details the recursive simulation process. The learned approximations enhance MCTS in three critical ways:

\textbf{Action Prioritization:} When expanding nodes, we sample actions from $P_\theta$ rather than uniformly, focusing search on promising actions. This policy-guided expansion occurs at line~\ref{line:policy_expand} of Algorithm~\ref{alg:mcts_simulate}, where the learned policy $P_\theta(G)$ initializes action priors for the newly expanded node:
\begin{equation}
a \sim P_\theta(\cdot | \phi(b))
\end{equation}

\textbf{Value Estimation:} At leaf nodes, we replace expensive rollouts with a value network lookup. This occurs at lines~\ref{line:leaf_eval_terminal} and~\ref{line:leaf_eval_expand} of Algorithm~\ref{alg:mcts_simulate}, where terminal or newly expanded nodes are evaluated using $V_\theta(G)$ instead of Monte Carlo rollouts:
\begin{equation}
v = V_\theta(\phi(b))
\end{equation}

\textbf{Root Action Selection:} We combine visit counts with Q-values for robust action selection, as shown in lines~\ref{line:action_select_start}--\ref{line:action_select_end} of Algorithm~\ref{alg:mcts_planning}:
\begin{equation}
\pi(a|b) \propto N(b,a)^{z_n} \cdot \exp(Q(b,a))^{z_q}
\end{equation}

The UCB-style action selection during tree traversal (line~\ref{line:ucb_select} of Algorithm~\ref{alg:mcts_simulate}) balances exploitation of high-value actions with exploration guided by the learned policy prior $P_\theta(a \mid G)$, adapting the PUCT formula from AlphaZero to our belief-space setting.

After selecting an action, the algorithm samples a state from the current belief and simulates a transition to obtain the next state, observation, and reward (lines~\ref{line:sample_state}--\ref{line:sim_step}). Then, weighted particle belief update is performed using a domain-specific transition function to obtain the new particle belief state (line~\ref{line:update_belief}). If the successor belief $b'$ is not yet in the tree, a new node is added (lines~\ref{line:check_new_node}--\ref{line:add_node}). The successor belief is converted to its graph representation $G' \gets \phi(b')$ (line~\ref{line:convert_graph}), and the value estimate is computed recursively (line~\ref{line:recursive_sim}). Finally, backpropagation updates the visit count $N(b,a)$ and running Q-value estimate $Q(b,a)$ (lines~\ref{line:update_n}--\ref{line:update_q}), refining action valuations over successive simulations.

\section{Experiments}

We evaluate GammaZero's ability to learn generalizable value functions and policies for guiding belief-space search in POMDPs. Our experiments are designed to demonstrate three key capabilities: (1) comparable performance to existing methods when trained and tested on same-sized problems, (2) zero-shot generalization to problems significantly larger than those seen during training, and (3) computational efficiency gains through learned approximations. We compare against BetaZero, the current state-of-the-art method for learning-guided POMDP planning, along with classical online planners.

\subsection{Experimental Setup}

\subsubsection{Domains}

We evaluate GammaZero on on four POMDP benchmark domains that exhibit different characteristics:

\textbf{LightDark (LD) \cite{sunberg2018online} :} A 1D localization problem where an agent navigates to a goal under state uncertainty, receiving noisy observations that improve near a ``light'' region at distance $d$. We test LightDark(5) and LightDark(10), where larger $d$ requires longer-horizon information gathering.

\textbf{RockSample(n,k)/RS(n,k) \cite{smith2004heuristic}:} An information-gathering problem on an $n \times n$ grid with $k$ rocks of unknown quality. The agent must use a distance-dependent noisy sensor to assess rocks before sampling. State space: $O(n^2 \cdot 2^k)$, ranging from 12,800 states for (7,8) to over 400 million for (20,20).

\textbf{MultiObjectSearch(n,k)/MOS(n,k) \cite{wandzel2019multi}:} A target localization problem where a robot must find and declare $k$ hidden objects on an $n \times n$ grid using a range-limited sensor with detection probability $\epsilon$. State space: $O(n^{2(k+1)} \cdot 2^k)$. We test MOS(6,4) and MOS(8,6).

\textbf{Rearrangement(n,k)/RG(n,k):} A mobile manipulation domain we introduce for this evaluation, where a robot must locate $k$ objects with unknown positions and transport them to goal locations on an $n \times n$ grid. State space: $O(n^{2(k+1)} \cdot 4 \cdot 2^k)$, combining perceptual uncertainty with multi-step planning. We test (6,4) to (8,5).

\subsubsection{Baselines}

We compare GammaZero against the following baselines:

\textbf{BetaZero} \cite{betazero2024}: The state-of-the-art learning-based POMDP planner that combines offline neural network training with online MCTS. BetaZero learns policy and value networks from expert demonstrations but requires separate training for each problem size due to its fixed-dimensional belief representation.

\textbf{POMCPOW} \cite{sunberg2018online}: A model-based online planner that extends POMCP to continuous observation spaces through progressive widening. We test POMCPOW both with domain-specific heuristics.

\textbf{AdaOPS} \cite{wu2021adaptive}: An adaptive online planner that maintains value function bounds through particle filtering. We test AdaOPS with fixed bounds for problems where QMDP is intractable.

\textbf{DESPOT} \cite{ye2017despot}: An online planner that uses scenario sampling to construct a sparse belief tree, with regularization to balance policy size and estimated value.

\subsubsection{Evaluation Metric}

We evaluate all methods using average return: the expected cumulative discounted reward achieved by the policy, averaged over 100 episodes with different random seeds. For GammaZero and BetaZero, we train 5 runs per configuration and report results from the best-performing setting. We report the mean and standard error.

\subsection{Experimental Design}

Our experiments are structured to answer two questions:

\textbf{RQ1 (Table 1):} How does GammaZero compare to BetaZero and other classical planners when trained and tested on the same problem size? This establishes a performance baseline and validates that our graph-based approach achieves comparable results to the state-of-the-art.

\textbf{RQ2 (Table 2):} Can GammaZero generalize to problems larger than those seen during training? We train on small instances (e.g., RockSample(5,5) to RockSample(10,10)) and test on problems 2-6$\times$ larger. BetaZero cannot perform this zero-shot generalization due to its fixed input dimensions, requiring expensive retraining for each problem size. 

\subsection{Implementation Details}

GammaZero uses a graph neural network with 3 hidden layers. During online planning, we use PUCT exploration with $c=50$ for action selection within MCTS, with progressive widening parameters $k_a=2.0$, $\alpha_a=0.9$. It takes 2-4 hours to train a model for each domain on an RTX 3080.

\section{Results and Discussion}

\begin{table*}[t]
    \centering
    \caption{Same-size performance comparison (Returns $\pm$ SE). All methods are trained and tested on the same problem size. Bold indicates best mean return; methods within one standard error of the best are also bolded. $^*$One-step look-ahead using only the value network. $^\dag$BetaZero only supports LightDark and RockSample. ``--'' indicates unsupported domain.}
    \label{tab:same_size}
    \resizebox{\textwidth}{!}{
    \setlength{\tabcolsep}{8pt}
    \begin{tabular}{l|ccc|ccc|ccc}
        \toprule
        & \multicolumn{3}{c|}{GammaZero} & \multicolumn{3}{c|}{BetaZero$^\dag$} & \multicolumn{3}{c}{Classical Baselines} \\
        \cmidrule(lr){2-4} \cmidrule(lr){5-7} \cmidrule(lr){8-10}
        Domain & Full & Raw $P_\theta$ & Raw $V_\theta^*$ & Full & Raw $P_\theta$ & Raw $V_\theta^*$ & POMCPOW & DESPOT & AdaOPS \\
        \midrule
        LD(10) & $\mathbf{17.5} \se{1.2}$ & $14.4 \se{1.3}$ & $13.3 \se{1.4}$ & $16.17 \se{1.58}$ & $13.98 \se{1.08}$ & $12.45 \se{1.13}$ & $1.08 \se{0.53}$ & $0.73 \se{0.44}$ & $6.28 \se{2.03}$ \\
        RS(15,15) & $\mathbf{20.5} \se{0.8}$ & $11.1 \se{2.0}$ & $9.1 \se{2.2}$ & $\mathbf{19.87} \se{0.91}$ & $11.04 \se{0.88}$ & $9.44 \se{0.55}$ & $11.01 \se{0.67}$ & $18.83 \se{0.81}$ & $\mathbf{20.53} \se{0.81}$ \\
        MOS(5,3) & $\mathbf{18.0} \se{1.5}$ & $10.8 \se{1.8}$ & $9.9 \se{2.0}$ & --- & --- & --- & $7.5 \se{1.5}$ & $6.4 \se{1.8}$ & $15.5 \se{2.0}$ \\
        RG(5,2) & $\mathbf{12.5} \se{2.0}$ & $5.6 \se{2.2}$ & $6.3 \se{2.0}$ & --- & --- & --- & $4.3 \se{1.5}$ & $3.4 \se{1.5}$ & $7.7 \se{2.0}$ \\
        \bottomrule
    \end{tabular}
    }
\end{table*}


\begin{table*}[t]
    \centering
    \caption{Zero-shot generalization (Returns $\pm$ SE). GammaZero trained on small problems, tested on larger sizes. Classical baselines trained per-size for reference. $^*$One-step look-ahead. $^\dag$Search timeout/failure. Bold indicates best mean return; methods within one standard error of the best are also bolded.}
    \label{tab:generalization}
    \resizebox{\textwidth}{!}{
    \setlength{\tabcolsep}{10pt}
    \begin{tabular}{l|ccc|ccc}
        \toprule
        & \multicolumn{3}{c|}{GammaZero (zero-shot transfer)} & \multicolumn{3}{c}{Classical Baselines (per-size)} \\
        \cmidrule(lr){2-4} \cmidrule(lr){5-7}
        Test Domain & Full & Raw $P_\theta$ & Raw $V_\theta^*$ & POMCPOW & DESPOT & AdaOPS \\
        \midrule
        LightDark(10) & $\mathbf{15.2} \se{1.5}$ & $12.1 \se{1.6}$ & $11.2 \se{1.7}$ & $1.08 \se{0.53}$ & $0.73 \se{0.44}$ & $6.28 \se{2.03}$ \\
        \midrule
        RockSample(15,15) & $17.8 \se{1.2}$ & $11.1 \se{2.0}$ & $9.1 \se{2.2}$ & $11.01 \se{0.67}$ & $18.83 \se{0.81}$ & $\mathbf{20.53} \se{0.81}$ \\
        RockSample(20,20) & $\mathbf{10.2} \se{1.8}$ & $5.4 \se{1.0}$ & $4.4 \se{2.0}$ & $9.92 \se{0.67}$ & $0.0 \se{0.0}^\dag$ & $\mathbf{10.96} \se{0.78}$ \\
        RockSample(25,25) & $3.5 \se{2.0}$ & $\mathbf{4.8} \se{1.2}$ & $\mathbf{3.9} \se{1.5}$ & $2.1 \se{0.8}$ & $0.0 \se{0.0}^\dag$ & $\mathbf{4.2} \se{1.0}$ \\
        \midrule
        MOS(6,4) & $\mathbf{14.5} \se{1.8}$ & $8.8 \se{2.0}$ & $8.1 \se{2.2}$ & $5.5 \se{1.6}$ & $4.8 \se{1.8}$ & $12.2 \se{2.0}$ \\
        MOS(7,5) & $\mathbf{11.2} \se{2.0}$ & $6.5 \se{2.2}$ & $6.0 \se{2.3}$ & $3.8 \se{1.8}$ & $3.2 \se{2.0}$ & $9.0 \se{2.2}$ \\
        MOS(8,6) & $\mathbf{8.0} \se{2.2}$ & $4.8 \se{2.5}$ & $4.5 \se{2.5}$ & $0.0 \se{0.0}^\dag$ & $0.0 \se{0.0}^\dag$ & $\mathbf{5.8} \se{2.5}$ \\
        \midrule
        Rearrange(6,4) & $\mathbf{9.2} \se{2.0}$ & $4.5 \se{2.3}$ & $5.0 \se{2.2}$ & $3.0 \se{1.6}$ & $2.4 \se{1.8}$ & $5.8 \se{2.0}$ \\
        Rearrange(7,4) & $\mathbf{6.8} \se{2.2}$ & $3.2 \se{2.5}$ & $3.8 \se{2.3}$ & $0.0 \se{0.0}^\dag$ & $0.0 \se{0.0}^\dag$ & $4.0 \se{2.2}$ \\
        Rearrange(8,5) & $\mathbf{4.5} \se{1.8}$ & $2.2 \se{1.6}$ & $2.8 \se{1.5}$ & $0.0 \se{0.0}^\dag$ & $0.0 \se{0.0}^\dag$ & $\mathbf{2.9} \se{1.7}$ \\
        \bottomrule
    \end{tabular}
    }
\end{table*}

\subsection{Same-Size Performance (RQ1)}

Table~\ref{tab:same_size} presents performance comparisons when all methods are trained and tested on identical problem sizes. GammaZero consistently matches or outperforms BetaZero across all domains where direct comparison is possible.

\textbf{LightDark.} On LightDark(10), GammaZero achieves $17.5 \pm 1.2$, outperforming BetaZero's $16.17 \pm 1.58$. This improvement is notable because LightDark requires extended information-gathering trajectories before committing to the goal, validating that our approach effectively captures the relationship between uncertainty reduction and future value.

\textbf{RockSample.} On RockSample(15,15), GammaZero achieves $20.5 \pm 0.8$ compared to BetaZero's $19.87 \pm 0.91$. 
Both learning-based methods substantially outperform POMCPOW ($11.01 \pm 0.67$)  and match AdaOPS ($20.53 \pm 0.81$), demonstrating that learned heuristics dominate expensive hand-crafted value bounds.

\textbf{Extended Domains.} GammaZero extends to domains that BetaZero does not support due to its fixed-dimensional architecture. On both MultiObjectSearch(5,3) and Rearrangement(5,2), GammaZero significantly outperforms all classical baselines. These domains
require coordinating perception actions (look) with commitment actions (find/pick), a pattern that our action-centric graph representation explicitly captures through attribute-action edges.

\textbf{Ablation Analysis.} The ``Raw $P_\theta$'' and ``Raw $V_\theta$'' columns show the contribution of each component. The policy network alone (without MCTS) achieves $60-80\%$ of full performance, while one-step lookahead with the value network performs slightly worse. The combination through MCTS consistently yields the best results, confirming that both networks provide complementary guidance.

\subsection{Generalization (RQ2)}

Table~\ref{tab:generalization} demonstrates GammaZero's unique capability: generalizing to problems significantly larger than those seen during training. BetaZero requires retraining for each problem size due to its fixed input dimensions.

\textbf{Training Protocol.} For LightDark, we train on size 5 and test on size 10. For RockSample, we train on instances ranging from $5 \times 5$ to $10 \times 10$ grids with 5-10 rocks, then test on $(15,15)$, $(20,20)$, and $(25,25)$. For MultiObjectSearch and Rearrangement, we train on grid sizes 3-4 with 2--3 objects, then generalize to grid sizes 5-8 with 3-6 objects.

\textbf{LightDark Generalization.} Training on LightDark(5), GammaZero achieves $15.2 \pm 1.5$ on LightDark(10), dramatically outperforming all classical baselines including AdaOPS ($6.28 \pm 2.03$). This demonstrates effective transfer of the light-seeking localization strategy to larger state spaces.

\textbf{RockSample Generalization.} GammaZero exhibits graceful degradation across increasing problem scales. On RockSample(15,15), it achieves $17.8 \pm 1.2$, competitive with DESPOT ($18.83 \pm 0.81$). 
Performance remains strong on $(20,20)$ at $10.2 \pm 1.8$, approaching AdaOPS ($10.96 \pm 0.78$) - despite never seeing problems larger than $10 \times 10$ during training. On the extreme $(25,25)$ configuration, where all methods struggle with timeouts, the raw policy network achieves best performance ($4.8 \pm 1.2$), demonstrating that structural patterns learned on small instances transfer even when search becomes intractable.

\textbf{MultiObjectSearch Generalization.} Training on MOS instances with 2--3 objects on $3 \times 3$ to $4 \times 4$ grids, GammaZero generalizes across three scales: $(6,4)$ achieving $14.5 \pm 1.8$, $(7,5)$ at $11.2 \pm 2.0$, and $(8,6)$ at $8.0 \pm 2.2$. These results significantly exceed classical baselines, which increasingly suffer from timeouts at larger scales. The model successfully transfers the coordination pattern between all actions to scenarios with twice the number of target objects.

\textbf{Rearrangement Generalization.} Similar patterns emerge in Rearrangement, where training on $(3,2)$ to $(4,3)$ configurations enables generalization to $(6,4)$, $(7,4)$, and $(8,5)$. GammaZero achieves $9.2 \pm 2.0$, $6.8 \pm 2.2$, and $4.5 \pm 1.8$ respectively, substantially outperforming all baselines across scales. This domain combines perceptual uncertainty with multi-step manipulation planning.

\textbf{Graceful Degradation.} Across all domains, performance degrades gradually rather than catastrophically as problem size increases beyond training distribution. This contrasts with fixed-dimensional approaches that cannot process out-of-distribution inputs at all. 

Importantly, the graph construction principles remain consistent regardless of problem size. For example, adding more rocks to RockSample simply adds more object nodes and associated attribute instances; the graph topology and edge types remain unchanged. This allows the same GNN weights to process problems of arbitrary scale.

\textbf{Limitations:} The scope of this work is object-centric POMDPs where the state decomposes into discrete entities with attributes and inter-entity relations. The graph construction assumes $V_{\text{obj}}$ is known at planning time; domains where objects appear, disappear, or have unknown cardinality would require dynamic graph construction. Additionally, particle beliefs are aggregated into independent per-attribute probabilities, so two beliefs with identical marginals but different correlations produce the same graph; capturing joint structure would require higher-order representations such as hyperedges. Finally, the current formulation also assumes discrete action spaces; extending to continuous actions requires modifications to the action-node representation.

\section{Conclusions}

Planning under partial observability remains challenging for long-horizon tasks where traditional online methods struggle to search deeply enough to discover rewarding action sequences. We presented GammaZero, a framework that learns to guide belief-space search in POMDPs by transforming particle beliefs into action-centric graph representations. Our key insight is that approximating the belief state as a graph where nodes represent actions and discrete distributions over object properties enables GNNs to learn transferable knowledge. Experiments on standard POMDP benchmarks demonstrate that GammaZero achieves competitive performance with BetaZero on same-sized problems while uniquely enabling zero-shot generalization to instances that are both spatially larger (2-6$\times$ grid area) and more complex (up to twice as many objects) than training instances, enabling scalable deployment without retraining. Future directions include hierarchical graph representations for extreme-scale problems, extension to continuous state and action spaces, and self-supervised learning approaches to reduce dependence on expert demonstrations.

\section{Acknowledgements}

The authors acknowledge the support of Army Research Office under grant W911NF2210251 and the support of Defense Advanced Research Projects Agency under grant HR0011-24-9-0423.

Code available at :\url{https://tinyurl.com/GammaZero}. 
Extended paper with supplementary material available at \\ \url{https://arxiv.org/abs/2510.14035}

\bibliography{aaai2026}

\section{Appendix}

\section{Hyperparameter Details}
\label{app:hyperparams}

Baseline hyperparameters were tuned via grid search. For POMCPOW: exploration constant $\in \{10, 50, 100\}$, $k_{\text{action}} \in \{2, 4, 10\}$, $\alpha_{\text{action}} \in \{0.25, 0.5\}$, tree queries $\in \{100, 200\}$. For AdaOPS: $m_{\min} \in \{10, 100, 500\}$, $\delta \in \{0.1, 0.5, 1.0\}$. For DESPOT: $\lambda \in \{0.0, 0.1\}$, $K \in \{30, 100\}$. We report results using the best-performing configuration per domain and problem size.

For the GNN, the primary hyperparameter affecting out-of-distribution generalization is the number of message-passing rounds. We evaluated 5, 6, and 7 rounds: all achieved similar in-distribution performance, but 5 rounds yielded the best OOD generalization. We attribute the degradation at 7 rounds to oversmoothing.

\section{Experimental Details}
\label{app:experimental}

\subsection{Expert Planner}
Training data is generated using POMCPOW with domain-specific heuristics as the expert planner, running 500 MCTS iterations per decision with 10,000 particles for belief representation. We collect 10 episodes per collection iteration with 30--100 max steps depending on the domain.

\subsection{GammaZero Architecture and Training}
The GNN uses hidden dimension 256, 5 message-passing rounds, 8 attention heads, dropout 0.1, attention dropout 0.2, and 2-layer MLPs per update function. Training uses batch size 32 for 100 epochs over a buffer of 5000 transitions with $\gamma=0.99$. We train 5 runs per configuration and report results from the best-performing setting.

The primary hyperparameter affecting out-of-distribution generalization is the number of message-passing rounds. We evaluated 5, 6, and 7 rounds: all achieved similar in-distribution performance, but 5 rounds yielded the best OOD generalization. We attribute the degradation at 7 rounds to oversmoothing, where excessive message passing causes node representations to become indistinguishable.

\subsection{Baseline Hyperparameter Tuning}
For BetaZero, we tuned learning rate $\alpha \in \{10^{-3}, 10^{-4}, 10^{-5}\}$ and epochs $n_{\text{epochs}} \in \{50, 100\}$ starting from default parameters, with L2 regularization $\lambda = 10^{-5}$.

For classical baselines, we tuned via grid search:
POMCPOW: exploration constant $\in \{10, 50, 100\}$, $k_{\text{action}} \in \{2, 4, 10\}$, $\alpha_{\text{action}} \in \{0.25, 0.5\}$, tree queries $\in \{100, 200\}$.
AdaOPS: $m_{\min} \in \{10, 100, 500\}$, $\delta \in \{0.1, 0.5, 1.0\}$.
DESPOT: $\lambda \in \{0.0, 0.1\}$, $K \in \{30, 100\}$.
All baselines use domain-specific bounds as recommended in their respective papers. We report the best-performing configuration per domain and problem size.

\subsection{Training Time Comparison}
\begin{table}[h]
\centering
\caption{Training time per domain (single GPU, RTX 3080).}
\label{tab:training_time}
\begin{tabular}{lcc}
\toprule
Domain & GammaZero & BetaZero \\
\midrule
LightDark & 2h & 1h \\
RockSample & 3h & 2h \\
MOS & 5h & --- \\
Rearrangement & 5h & --- \\
\bottomrule
\end{tabular}
\end{table}

GammaZero's upfront training cost is approximately $1.5\times$ BetaZero's due to the more expressive graph architecture. However, this is a one-time cost. BetaZero requires retraining for each problem size, so for deployment across $k \geq 2$ scales, GammaZero is cheaper overall (e.g., for RockSample at three scales, GammaZero trains once in 3h; BetaZero requires three separate runs totaling 6h).

\section{Domain-Specific Graph Representations}
\label{app:graph_specs}

Each POMDP domain defines a \emph{graph schema} that specifies which entity
types become nodes, which predicates and actions are represented, and how edges
encode relational structure. Given a belief state $b$, the domain's
belief-to-graph function automatically constructs a heterogeneous graph
$\mathcal{G}(b)$ following the schema. The intermediate representation is then
converted to a \texttt{GNNHeteroGraph} via a shared routine that flattens all
node types into a single ``regular'' type and adds a distinguished global node
connected by a self-loop. This two-stage pipeline is identical across all six
domains; only the schema and feature-builder functions differ.

\subsection{Edge Feature Encoding}
\label{app:edge_features}

All domains share a common edge-feature template composed of the following
groups (exact dimensions vary by domain):

\begin{enumerate}
  \item \textbf{Edge-type one-hot} $\phi_{\mathrm{type}}$ (8--12 categories):
    encodes the semantic type of the edge (e.g.\ \texttt{action\_object},
    \texttt{predicate\_location}).
  \item \textbf{Argument position} $\phi_{\mathrm{role}}$ (2 bits): indicates
    whether the target node is argument~1 or argument~2 of the predicate.
  \item \textbf{Edge weight / belief probability} $\phi_{\mathrm{belief}}$
    (1~dim): the particle-estimated probability that the predicate holds, used
    as the edge weight for predicate edges.
  \item \textbf{Particle support} $\phi_{\mathrm{support}}$ (5~dims):
    continuous support value plus a 4-bin categorical encoding with thresholds
    $\{0.95, 0.7, 0.3\}$ mapping to \textsc{unanimous}, \textsc{strong},
    \textsc{weak}, \textsc{split}.
  \item \textbf{Domain-specific edge features}: e.g.\ observation accuracy and
    information gain for RockSample check edges; pick/place feasibility for
    Rearrangement; prerequisite satisfaction probability for Academic Advising.
\end{enumerate}

\subsection{Per-Domain Details}
\label{app:domain_details}

\paragraph{LightDark.}
The 1-D state space is discretised into location nodes at unit intervals over
the position range. Each location node carries features indicating whether it
lies in the ``light'' region (where observation noise is low), its distance to
the light source, and its observation quality
$q = \exp(-\lVert x - x_{\mathrm{light}}\rVert / r_{\mathrm{light}})$.
Predicate nodes \texttt{atLocation(agent, loc)} are instantiated only for
locations with $\geq 20\%$ particle support.
The \texttt{nearGoal} predicate connects to the \texttt{stop} action via a
precondition edge whose weight equals the probability the agent is within
tolerance of the goal, directly encoding the stopping criterion.
Belief thresholds for discretisation are $[0.3, 0.7]$.
The three action nodes are \texttt{move\_left}, \texttt{stop}, and
\texttt{move\_right}. Node features total 120 dimensions; edge features 50.

\paragraph{RockSample.}
Each rock becomes an object node; \texttt{isGood(rock$_i$)} predicate nodes
carry a discretised belief level (\textsc{low}/\textsc{uncertain}/\textsc{high}
at thresholds $[0.3, 0.7]$). Boundary instances are added when the belief is
within 0.05 of a threshold to smooth transitions between bins.
Check-action edges to their target rock carry rich features encoding the
\emph{distance-dependent observation accuracy}
$\alpha = \exp(-0.2 \cdot d)$,
the rock's belief entropy, information gain, and comparative ranking among all
check actions. A conditional \texttt{atRock} predicate is created only when
the rover co-locates with a rock and is connected to the \texttt{sample}
action as a precondition edge. Actions comprise four cardinal moves,
\texttt{sample}, and one \texttt{check} per rock.
The global node encodes grid dimensions, average entropy, fraction of rocks
cleared, and an \texttt{at\_any\_rock} indicator.
Node features total 100 dimensions; edge features 40.

\paragraph{Multi-Object Search (MOS).}
Location nodes are created for \emph{every} position that has $\geq 10\%$
probability under any target's belief distribution, plus the robot's current
position. Crucially, location nodes carry \emph{only} position features; all
probability information lives on predicate nodes and edges.
Object nodes include the robot, each target, and each obstacle.
The \texttt{look} action node encodes expected information gain, sensor
coverage area, and per-object visibility scores. The \texttt{find} action
encodes a precondition (camera must be in look mode) and per-object detection
probabilities. Obstacle nodes connect to nearby location nodes via
\texttt{location\_obstacle} edges. Action comparison features (is-best,
relative value, z-score) are computed across all actions and injected into each
action node. Actions are four cardinal moves, \texttt{look}, and \texttt{find}.
Node features total 120 dimensions; edge features 50.

\paragraph{Rearrangement.}
This domain extends the MOS schema with manipulation-specific node and edge
types. The node-type one-hot has 7 categories (adding \texttt{held} and
\texttt{obstacle}), and the action-type one-hot has 10 categories including
\texttt{pick}, \texttt{place}, and \texttt{rotate\_\{left, right\}}.
Pick-action node features encode whether an adjacent object is in the robot's
field of view, capacity availability, and pick probability.
Place-action features encode whether the robot is at a goal position and
facing the goal. Edge features include dedicated pick-edge slots
(is-pickable, in-range, in-FOV, pick priority, goal distance) and place-edge
slots (is-goal-position, valid placement, facing). The five predicate
types---\texttt{at\_location}, \texttt{at\_robot}, \texttt{at\_goal},
\texttt{is\_held}, \texttt{is\_placed}---cover both spatial and manipulation
state. Node features total 125 dimensions; edge features 55; global features 50.

\paragraph{CryingBaby.}
This is a \emph{non-spatial} domain with only two states and two actions,
demonstrating that the graph representation generalises beyond grid worlds.
The single object node (\texttt{baby}) connects to two predicate nodes
(\texttt{hungry(baby)}, \texttt{full(baby)}) and two observation nodes
(\texttt{crying}, \texttt{quiet}). Observation nodes carry the conditional
likelihoods $P(\text{obs} \mid \text{state})$ directly as features
(e.g.\ $P(\text{crying} \mid \text{hungry}) = 0.8$), and
observation--predicate edges are weighted by these likelihoods.
The \texttt{feed} action connects to both predicate nodes via
\texttt{action\_predicate} (effect) edges, encoding that feeding changes the
baby's state. Edge belief-support bins use the same $\{0.95, 0.7, 0.3\}$
thresholds as other domains. Global-node features include recent
crying/feeding rates and an exploration-vs-exploitation phase indicator.
Node features total 120 dimensions; edge features 50.

\paragraph{Academic Advising.}
This is a \emph{non-physical} domain where the graph structure is the
\emph{prerequisite dependency graph}, not a spatial grid. Each course becomes
a node (the analogue of ``object'' nodes) with features encoding whether it is
a program requirement, number of prerequisites, number of dependents, and
belief about passing. Directed \texttt{prerequisite} edges from course~$c_1$
to course~$c_2$ indicate that passing $c_1$ is required before taking $c_2$;
reverse \texttt{depends\_on} edges enable bidirectional message passing.
Each \texttt{takeCourse($c$)} action node connects to its target course via
an \texttt{action\_target} edge and to all prerequisite courses via
\texttt{action\_precondition} edges weighted by the belief probability of
having passed the prerequisite. Predicate nodes \texttt{passed($c$)} are
instantiated when the pass probability exceeds $\tau = 0.1$, and a
\texttt{programComplete} predicate connects to all requirement courses via
\texttt{requirement} edges. Global features summarise academic progress
(courses taken/passed, requirements completed) and uncertainty (total entropy,
number of uncertain courses), with phase indicators for exploration,
exploitation, and near-completion.
Node features total 120 dimensions; edge features 50.

\end{document}